\documentclass[journal]{elsarticle}
\pdfoutput=1
\usepackage{mathtools}
\usepackage{subfig}
\usepackage{graphicx}
\usepackage{times}
\usepackage{graphicx}
\usepackage{multirow}
\usepackage{breqn}
\usepackage{tabularx}
\usepackage{epstopdf}

\newcolumntype{Y}{>{\centering\arraybackslash}X}

\usepackage{epstopdf}
\usepackage{amssymb}
 \usepackage{algorithm, algorithmic}
\usepackage{enumitem}
\usepackage{booktabs,siunitx}

\usepackage{array}
\usepackage{ifpdf}
\usepackage{mathtools}
\usepackage{lipsum}
\usepackage{amsmath, amsfonts, pifont, float, color, url}
\usepackage{graphicx}
\usepackage[draft]{hyperref}
\newcolumntype{x}[1]{%
>{\centering\hspace{0pt}}m{#1}}%

\usepackage{array}

\usepackage{multirow}

\begin{document}
\thispagestyle{empty}	
\title{Contextual Audio-Visual Switching For Speech Enhancement in Real-World Environments}
\author{Ahsan Adeel, Mandar Gogate, Amir Hussain}

\address{University of Stirling, Stirling FK9 4LA, United Kingdom}


\begin{abstract}
Human speech processing is inherently multimodal, where visual cues (lip movements) help to better understand the speech in noise. Lip-reading driven speech enhancement significantly outperforms benchmark audio-only approaches at low signal-to-noise ratios (SNRs). However, at high SNRs or low levels of background noise, visual cues become fairly less effective for speech enhancement, and audio-only cues work well enough. Therefore, a more optimal, context-aware audio-visual (AV) system is required, that contextually utilises both visual and noisy audio features and effectively accounts for different noisy conditions. In this paper, we introduce a novel contextual AV switching component that contextually exploits AV cues with respect to different operating conditions to estimate clean audio, without requiring any SNR estimation. The switching module switches between visual-only (V-only), audio-only (A-only), and both audio-visual cues at low, high and moderate SNR levels, respectively. The contextual AV switching component is developed by integrating a convolutional neural network (CNN) and long-short-term memory (LSTM) network. For testing, the estimated clean audio features are utilised by the developed novel enhanced visually derived Wiener filter (EVWF) for clean audio power spectrum estimation. The contextual AV speech enhancement method is evaluated under dynamic real-world scenarios (cafe, street, BUS, pedestrian) at different SNR levels (ranging from low to high SNRs) using benchmark Grid and ChiME3 corpora. For objective testing, perceptual evaluation of speech quality (PESQ) is used to evaluate the quality of the restored speech. For subjective testing, the standard mean-opinion-score (MOS) method is used. The critical analysis and comparative study demonstrate the outperformance of proposed contextual AV approach, over A-only, V-only, spectral subtraction (SS), and log-minimum mean square error (LMMSE) based speech enhancement methods at both low and high SNRs, revealing its capability to tackle spectro-temporal variation in any real-world noisy condition. Simulation results also validate the phenomenon of less effective visual cues at high SNRs, less effective audio cues at low SNRs, and complementary audio and visual cues strength. Lastly, the benefit of using visual cues at low SNRs is demonstrated using colour spectrogram where visual cues better recovered the speech components at specific time-frequency units as compared to A-only cues.  
\end{abstract}
\maketitle


\section{Introduction}
The ability to partially understand speech by looking at the speaker's lips is known as lip reading. Lip reading is helpful to all sighted people, both with normal and impaired hearing. It improves speech intelligibility in noisy environment, especially, when audio only perception is compared with AV perception. Lip reading confers benefits because the visible articulators, such as lips, teeth, and tongue are correlated with the resonance of the vocal tract. The correlation between the visible properties of the articulatory organs and speech reception has been previously shown in numerous behavioural studies \cite{sumby1954visual}\cite{summerfield1979use}\cite{mcgurk1976hearing}\cite{patterson2003two}. Early studies such as \cite{middelweerd1987effect} \cite{macleod1990procedure} have shown that lip reading could help tolerate an extra 4-6 dB of noise, where one decibel SNR gain is equivalent to 10-15\% better intelligibility. In high-noise environments (and also in reverberation), the auditory representation in mid to high frequencies is often degraded \cite{summerfield1979use}. In such cases, people with hearing impairment require the SNR to be improved by about 1 dB for every 10 dB of hearing loss, if they are to perceive speech as accurately as listeners with normal hearing \cite{duquesnoy1983effect}.


In the literature, extensive research has been carried out to develop multimodal speech processing methods, which establishes the importance of multimodal information in speech processing \cite{katsaggelos2015audiovisual}. Researchers have proposed novel visual feature extraction methods \cite{chen2013pattern}\cite{waibel1990readings}\cite{bundy1984linear}\cite{zhou2014review}\cite{sui2012discrimination}, fusion approaches (early integration \cite{nefian2002dynamic}, late integration \cite{snoek2005early}, hybrid integration \cite{wu2006multi}), multi-modal datasets \cite{cooke2006audio}\cite{patterson2002cuave}, and fusion techniques \cite{wu2006multi}\cite{iyengar2003discriminative}\cite{noulas2006detection}. The multimodal audiovisual speech processing methods have shown significant performance improvement in ASR, speech enhancement and speech separation \cite{berthommier2004phonetically}
\cite{chung2016lip} \cite{wu2016multi} \cite{noda2015audio} \cite{gogate2018dnn}. Recently, the authors in \cite{hou2018audio} developed an audio-visual deep CNN (AVDCNN) speech enhancement model that integrates audio and visual cues into a unified network model. The proposed AVDCNN approach is structured as an audio-visual encoder-decoder where both audio and visual cues are processed using two separate CNNs, and later fused into a joint network to generate enhanced speech. However, the proposed AVDCNN approach relies only on deep CNN models. For testing, the authors used self-prepared dataset that contained video recordings of 320 utterances of Mandarin sentences spoken by a native speaker (only one speaker). In addition, the used noises include usual car engine, baby crying, pure music, music with lyrics, siren, one background talker (1T), two background talkers (2T), and three background talkers (3T). In summary, for testing only one Speaker (with 40 clean utterances mixed with 10 different noise types at 5 dB, 0 dB, and -5 dB SIRs) was considered. 

In contrast to \cite{hou2018audio}, the proposed speech enhancement framework leverages the complementary strengths of both deep learning and analytical acoustic modelling (filtering based approach). In addition, the introduced contextual AV component effectively accounts for different noisy conditions by contextually utilising both visual and noisy audio features. It is believed that our brain works in a same way by contextually exploiting and integrating multimodal cues such as lip-reading, facial expressions, body language, gestures, audio, and textual information to further enhance the speech quality and intelligibility. However, limited work has been conducted to contextually exploit multimodal AV switching for speech enhancement applications in different noisy scenarios. 


The rest of the paper is organised as follows: Section 2 presents the proposed speech enhancement framework including contextual audio-visual switching and EVWF. Section 3 presents the employed AV dataset and audiovisual feature extraction methodology. In Section 4, comparative experimental results are presented. Finally, Section 5 concludes this work with some future research directions.
 \begin{figure}[!t]
 	\centering
 	\includegraphics[trim=0cm 0cm 0cm 0cm, clip=true, width=0.85\textwidth]{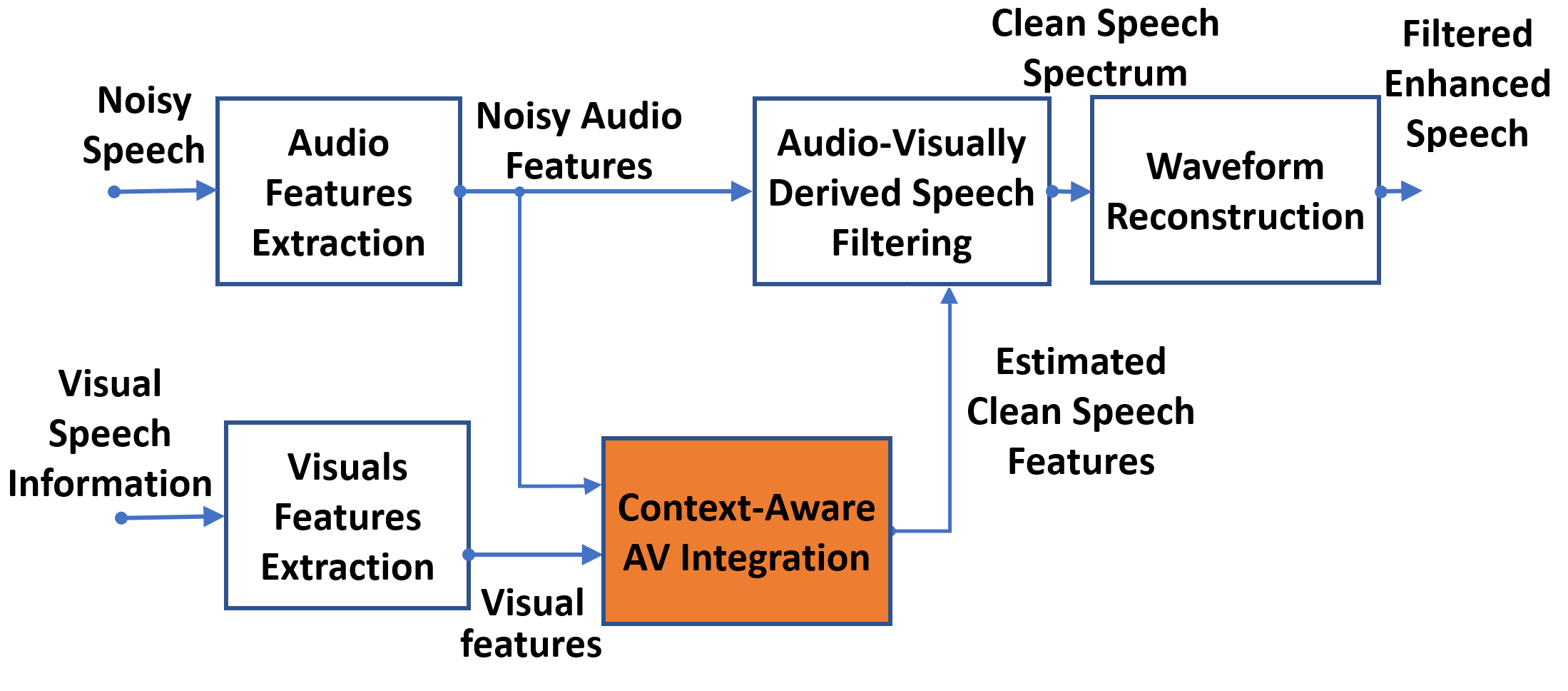}
 	\caption{Contextual audio-Visual switching for speech enhancement. The system contextually exploits audio and visual cues to estimate clean audio features. Afterwards, the estimated clean audio features are feeded into the proposed enhanced visually derived Wiener filter for the estimation of clean speech spectrum.}
 	\label{fig:Overview}
 \end{figure}
 \begin{figure*} [!t]
 	\centering
 	\includegraphics[trim=0cm 0cm 0cm 0cm, clip=true, width=1\textwidth]{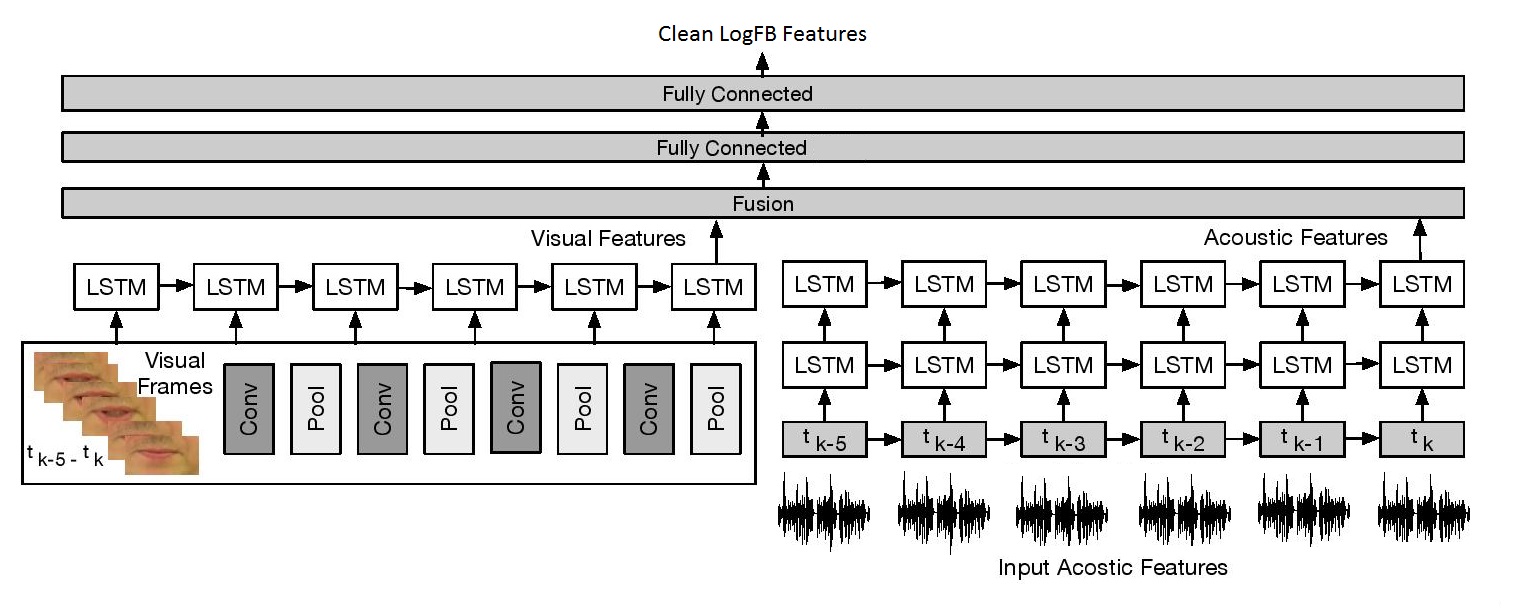}
 	\caption{Contextual audio-visual switching module: An integration of CLSTM and LSTM with 5 Prior Audio and Visual Frames (taking into account the current AV frame $t_k$ as well as the temporal information of previous 5 AV frames $t_{k-1}$, $t_{k-2}$, $t_{k-3}$, $t_{k-4}$, $t_{k-5}$)}
 	\label{fig:Framework}
 \end{figure*}

\section{Speech Enhancement Framework} 
A schematic block diagram of the proposed context-aware speech enhancement framework is depicted in Figure \ref{fig:Overview}. The proposed architecture comprises an audio and video feature extraction module, a context-aware AV switching component, a speech filtering component, and a waveform reconstruction module. The contextual AV switching component exploits both audio and visual features to learn the correlation between input (noisy audio and visual features) and output (clean speech) in different noisy scenarios. The designed context-aware AV switching component is composed of two deep learning architectures, an LSTM and CLSTM. The LSTM and CLSTM driven estimated clean audio features are then feeded into the designed EVWF to calculate the optimal Wiener filter, which is then applied to the magnitude spectrum of the noisy input audio signal, followed by the waveform reconstruction process. More details are comprehensively presented in Section 2.1 and Section 2.2.

\subsection{Contextual Audio-Visual switching} 
The proposed deep learning driven context-aware AV switching architecture is depicted in Figure \ref{fig:Framework} that leverages  the complementary strengths of CNN and LSTM. The term context-aware signifies the learning and adaptable capabilities of the proposed AV switching component, that helps exploiting the multimodal cues contextually to better deal with all possible speech scenarios (e.g. low, medium, high background noises, different environments (home, restaurant, social gathering etc)). The CLSTM network for visual cues processing consists of input layer, four Convolution layers (with 16, 32, 64, and 128 filters of size 3 $\times$ 5 respectively), four Max Pooling layers of size 2 $\times$ 2, and one LSTM layer (100 cells). The LSTM network for audio cues processing consists two LSTM layer. Visual features of time instance $t_k,t_{k-1},...,t_{k-5}$ (\textit{k} is the current time instance and 5 is the number of prior visual frames) were feeded into the  CLSTM model to extract optimised features which were then exploited by 2 LSTM layers. Audio features of time instance $t_k,t_{k-1},...,t_{k-5}$ were feeded into the LSTM model with two LSTM layers. The first LSTM layer has 250 cells, which encoded the input and passed its hidden state to the second LSTM layer, which has 300 cells. Finally, the optimised latent features from both CLSTM and LSTM were fused using two dense layers. The architecture was trained with the objective to minimise the mean squared error (MSE) between the predicted and the actual audio features. The MSE (1) between the estimated audio logFB features and clean audio features was minimised using stochastic gradient decent algorithm and RMSProp optimiser. RMSprop is an adaptive learning rate optimiser which divides the learning rate by moving average of the magnitudes of recent gradients to make learning more efficient. Moreover, to reduce the overfitting, dropout (0.20)  was applied after every LSTM layer. The MSE cost function $C(a_{estimated},a_{clean})$ can be written as:

\begin{equation}
C(a_{\text{estimated}},a_{\text{clean}}) = \sum_{i=1}^{n} 0.5  (a_{\text{estimated}}(i)-a_{\text{clean}}(i))^{2}
\end{equation}

where $a_{estimated}$ and $a_{clean}$ are the estimated and clean audio features respectively.

\begin{figure}
	\centering 
	\subfloat[State-of-the-art VWF \cite{almajai2011visually}. ]{\includegraphics[width=0.5\textwidth]{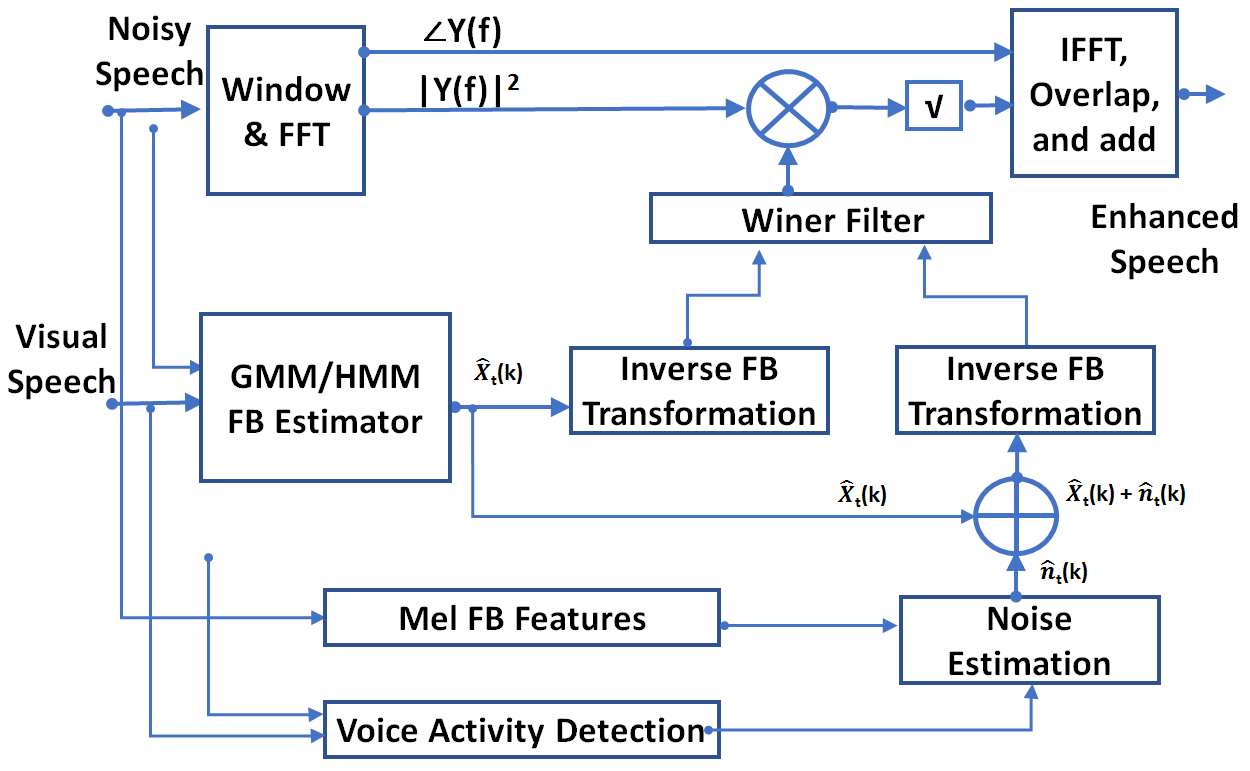}}
	\subfloat[Proposed enhanced VWF (EVWF)]{\includegraphics[width=0.5\textwidth]{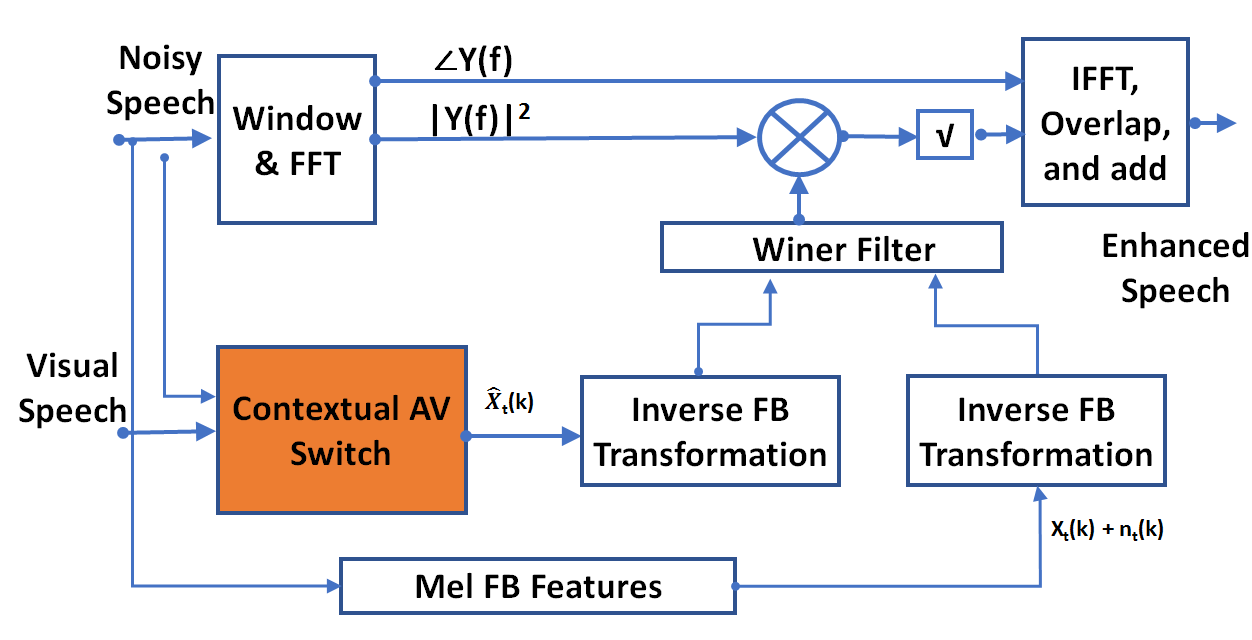}} \\

	\caption{Visually-derived Wiener filtering (VWF)} 
	\label{fig:VWF} 
\end{figure}

\subsection{Enhanced Visually Derived Wiener Filter}
The EVWF transforms the estimated low dimensional clean audio features into high dimensional clean audio power spectrum using inverse filter-bank (FB) transformation and calculates Wiener filter. The Wiener filter is then applied to the magnitude spectrum of the noisy input audio signal, followed by the inverse fast Fourier transform (IFFT), overlap, and combining processes to produce enhanced magnitude spectrum. The state-of-the-art VWF and designed EVWF are depicted in Figure \ref{fig:VWF} (a) and (b) respectively. The authors in \cite{almajai2011visually} presented a hidden Markov model-Gaussian mixture model (HMM/GMM) based two-level state-of-the-art VWF for speech enhancement. However, the use of HMM/GMM and cubic spline interpolation for clean audio features estimation and low to high dimensional transformation are not optimal choices. This is due to HMM/GMM models limited generalisation capability and poor audio power spectrum estimation of cubic spline interpolation method that fails to estimate the missing power spectral values. In contrast, the proposed EVWF exploited LSTM based FB estimation and an inverse FB transformation for audio power spectrum estimation which addressed both power spectrum estimation and generalisation issues in VWF. In addition, the proposed EVWF replaced the need of voice activity detector and noise estimation. 

The frequency domain Wiener Filter is defined as:

\begin{eqnarray}
W(\gamma)=\frac{\psi\hat{a}(\gamma)}{\psi_a (\gamma)}
\end{eqnarray}

where $\psi_a (\gamma)$ is the noisy audio power spectrum (i.e. clean power spectrum + noisy power spectrum) and $\psi\hat{a}(\gamma)$ is the clean audio power spectrum. The calculation of the noisy audio power spectrum is fairly easy because of the available noisy audio vector. However, the calculation of the clean audio power spectrum is challenging which restricts the use of Wiener filter widely. Hence, for successful Wiener filtering, it is necessary to acquire the clean audio power spectrum. In this paper, the clean audio power spectrum is calculated using deep learning based lip reading driven speech model. 

%
%


The FB domain Wiener filter ($\hat{W}_t^{FB}(k)$) is given as \cite{almajai2011visually}:

\begin{equation}
\hat{W}_t^{FB}(k)=\frac{\hat{x}_t(k)}{\hat{x}_t(k) + \hat{n}_t(k)}
\end{equation}


where $\hat{x}_t(k)$ is the FB domain lip-reading driven approximated clean audio feature and $\hat{n}_t(k)$ is the FB domain noise signal. The subscripts \textit{k} and \textit{t} represents the \textit{$k^{th}$} channel and \textit{$t^{th}$} audio frame. 

%

The lip-reading driven approximated clean audio feature vector $\hat{x}_t(k)$ is a low dimensional vector. For high dimensional Wiener filter calculation, it is necessary to transform the estimated low dimensional FB domain audio coefficients into a high dimensional power spectral domain. It is to be noted that the approximated clean audio and noise features ($\hat{x}_t(k) + \hat{n}_t(k)$) are replaced with the noisy audio (a combination of real clean speech ($x_t(k))$ and noise $(n_t(k)$)). The low dimension to high dimension transformation can be written as:



%

\begin{equation}
\hat{W}_     {t_{[N_l, M]}}     ^{FB}(k)=\frac{  \hat{x}_ {t(k)_{[N_l, M]}}    }{x_ {t(k)_{[N_l, M]}}  + n_ {t(k)_{[N_l, M]}} }
\end{equation}

\begin{equation}
\hat{W}_     {t_{[N_h, M]}}     ^{FB}(k)=\frac{  \hat{x}_ {t(k)_{[N_h, M]}}    }{x_ {t(k)_{[N_h, M]}}  + n_ {t(k)_{[N_h, M]}} }
\end{equation}

Where $N_l$ and $N_h$ are the low and high dimensional audio features respectively, and M is the number of audio frames. The transformation from (4) to (5) is very crucial to the performance of filtering. Therefore, the authors in \cite{almajai2011visually} proposed the use of cubic spline interpolation method to determine the missing spectral values. In contrast, this article proposes the use of inverse FB transformation and is calculated as follows:

\begin{equation}
\hat{x}_ {t(k)_{[N_h, M]}} = \hat{x}_ {t(k)_{[N_l, M]}} * \alpha_x
\end{equation}

\begin{equation}
n_ {t(k)_{[N_h, M]}} = n_ {t(k)_{[N_l, M]}} * \alpha_n
\end{equation}

\begin{equation}
\alpha_x = \alpha_n = (\phi_m(k)^T \phi_m(k))^{-1}\phi_m(k)^T
\end{equation}

$\phi_m(k)$=\[ \begin{cases} 
0 & k<f_{b_{mf-1}} \\
\frac{k-f_{b_{mf-1}}}{f_{b_{mf}}-f_{b_{mf-1}}} & f_{b_{mf-1}}\leq k\leq f_{b_{mf}} \\
\frac{f_{b_{mf-1}}-k}{f_{b_{mf+1}}+f_{b_{mf}}} & f_{b_{mf}}\leq k\leq f_{b_{mf+1}} \\
0 & k> f_{b_{mf+1}} 
\end{cases}
\]

where $f_{b_{mf}}$ are the boundary points of the filters and corresponds to the $k^{th}$ coefficient of the \textit{k}-points DFT. The boundary points $f_{b_{mf}}$ are calculated using:

\begin{equation}
f_{b_{mf}}=(\frac{K}{F_{samp}}.f_{cm_{f}})
\end{equation} 

where $f_{cm_{f}}$ is the mel scale frequency



After substitutions, the obtained \textit{k}-bin Wiener filter (5) is applied to the magnitude spectrum of the noisy audio signal (i.e. $|Y_t(k)|$) to estimate the enhanced magnitude audio spectrum ($|\hat{X}_ {t(k)_{[N_H, M]}} |$). The enhanced magnitude audio spectrum is given as:

\begin{eqnarray}
|\hat{X}_ {t(k)}|=|Y_{t(k)}| |	\hat{W}_     {t_{[N_h, M]}} |
\end{eqnarray}

The acquired time-domain enhanced speech signal (10) is then followed by the IFFT, overlap, and combining processes.

\section{AudioVisual Corpus \& Feature Extraction}
In this section we present the developed AV ChiME3 corpus and our feature extraction pipeline. 
\subsection{Dataset}
The AV ChiME3 corpus is developed by mixing the clean Grid videos \cite{cooke2006audio} with the ChiME3 noises \cite{barker2015third} (cafe, street junction, public transport (BUS), pedestrian area) for SNRs ranging from -12 to 12dB. The preprocessing includes sentence alignment and incorporation of prior visual frames. The sentence alignment is performed to remove the silence time from the video and prevent model from learning redundant or insignificant information. Prior multiple visual frames are used to incorporate temporal information to improve mapping between visual and audio features. The Grid corpus comprised of 34 speakers each speaker reciting 1000 sentences. Out of 34 speakers, a subset of 5 speakers is selected (two white females, two white males, and one black male) with total 900 command sentences each. The subset fairly ensures the speaker independence criteria. A summary of the acquired visual dataset is presented in Table 1, where the full and aligned sentences, total number of sentences, used sentences, and removed sentences are clearly defined. 
\begin{table}
	\centering 
	\caption{Used Grid Corpus Sentences}  
	\begin{tabular}{|c|c|c|c|c|c|c|}    
		\cline{4-7}                      
		\multicolumn{3}{c}{} &  \multicolumn{2}{|c|}{Full Sentences} &  \multicolumn{2}{|c|}{Aligned Sentences}\\
		\hline
		Speaker ID & Grid ID & No. of Sentences &  Removed &  Used & Removed & Used\\
		\hline
		\hline	
		Speaker 1 & S1 & 1000 & 11 & 989 & 11 & 989\\
		Speaker 2 & S15 & 1000 & 164 & 836 & 164 & 836\\
		Speaker 3 & S26 & 1000 & 16 & 984 & 71 & 929\\
		Speaker 4 & S6 & 1000 & 9 & 991 & 9 & 991\\
		Speaker 5 & S7 & 1000 & 11 & 989 & 11 & 989\\		
		\hline	 		
	\end{tabular}
	
	\label{table:featextract-sentnum}    
\end{table}

\begin{table}[]
	\centering
	\caption{Summary of the Train, Test, and Validation Sentences}
	\begin{tabular}{|c|c|c|c|c|c|c|}
		
		\hline
		Speakers & Train & Validation & Test & Total \\
		\hline
		\hline
		
		1        & 692   & 99         & 198  & 989   \\
		2        & 585   & 84         & 167  & 836   \\
		3        & 650   & 93         & 186  & 929   \\
		4        & 693   & 99         & 199  & 991   \\
		5        & 692   & 99         & 198  & 989   \\
		All      & 3312  & 474        & 948  & 4734 \\
		\hline
		
	\end{tabular}
\end{table} 
\subsection{Audio feature extraction}
The audio features are extracted using widely used log-FB vectors. For log-FB vectors calculation, the input audio signal is sampled at 50kHz and segmented into \textit{N} 16ms frames with 800 samples per frame and 62.5\% increment rate. Afterwards, a hamming window and Fourier transformation is applied to produce 2048-bin power spectrum. Finally, a 23-dimensional log-FB is applied, followed by the logarithmic compression to produce 23-D log-FB signal. 
\subsection{Visual feature extraction}
The visual features are extracted from the Grid Corpus videos recorded at 25 fps using a 2D-DCT based standard and widely used visual feature extraction method. Firstly, the video files are processed to extract a sequence of individual frames. Secondly, a Viola-Jones lip detector \cite{viola2001rapid} is used to identify lip-region by defining the Region-of-Interest (ROI) in terms of bounding box. Object detection is performed using Haar feature-based cascade classifiers. The method is based on machine learning where cascade function is trained with positive and negative images. Finally, the object tracker \cite{ross2008incremental} is used to track the lip regions across the sequence of frames. The visual extraction procedure produced a set of corner points for each frame, where lip regions are then extracted by cropping the raw image. In addition, to ensure good lip tracking, each sentence is manually validated by inspecting few frames from each sentence. The aim of manual validation is to delete those sentences in which lip regions are not correctly identified \cite{abel2016data}. Lip tracking optimisation lies outside the scope of the present work. The development of a full autonomous system and its testing on challenging datasets is in progress.      
\begin{table}
	\centering 
	\caption{LSTM (V-only model) training and testing accuracy comparison for different prior visual frames}  
	\begin{tabular}{|c|c|c|}    
		\cline{2-3}                      
		\multicolumn{1}{c}{} &  \multicolumn{2}{|c|}{LSTM - V-only} \\
		\hline
		Visual Frames &  $MSE_{train}$ & $MSE_{test}$ \\
		\hline
		\hline	
		1  & 0.092 & 0.104 \\
		2   & 0.087 & 0.097 \\
		4   & 0.073 & 0.085 \\
		8  & 0.066 & 0.082 \\
		14  & 0.0024 & 0.0020 \\		
		18  & 0.0016 & 0.019 \\		
		\hline	 		
	\end{tabular}
	\label{table:featextract-sentnum}  
\end{table}
\section{Experimental Results}
\begin{figure} 
	\centering
	\includegraphics[trim=0cm 0cm 0cm 0cm, clip=true, width=0.65\textwidth]{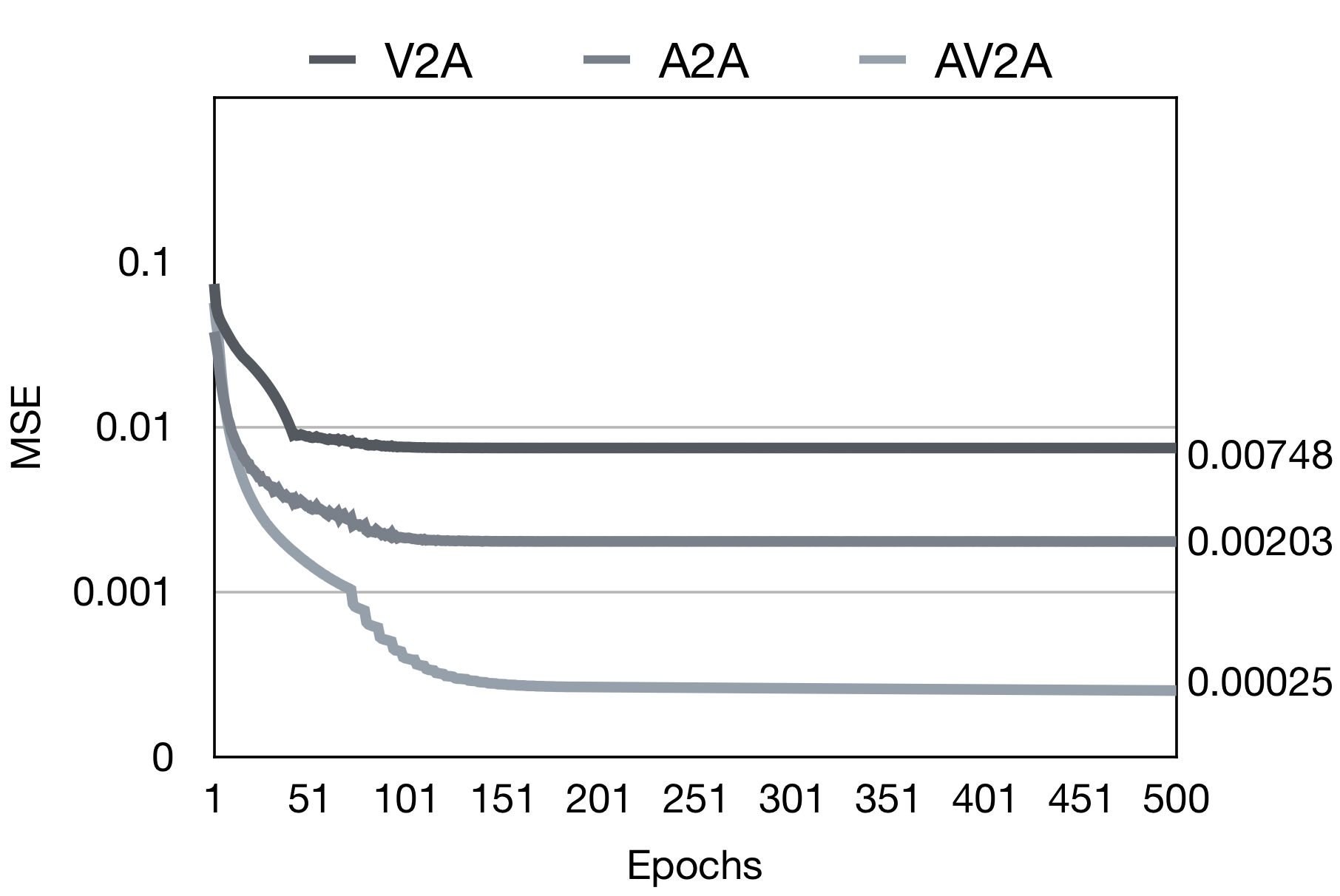}
	\caption{Validation Results for A-only (A2A), V-only (V2A), and contextual AV (AV2A) mapping with 14 prior AV frames - All Speakers. The figure presents an overall behaviour of A2A, V2A, and AV2A mapping. It is to be noted that contextual AV mapping outperforms both A-only and V-only mappings.}
	\label{fig:MSEComparison}
\end{figure} 

\begin{figure} 
	\centering
	\includegraphics[trim=0cm 0cm 0cm 0cm, clip=true, width=0.65\textwidth]{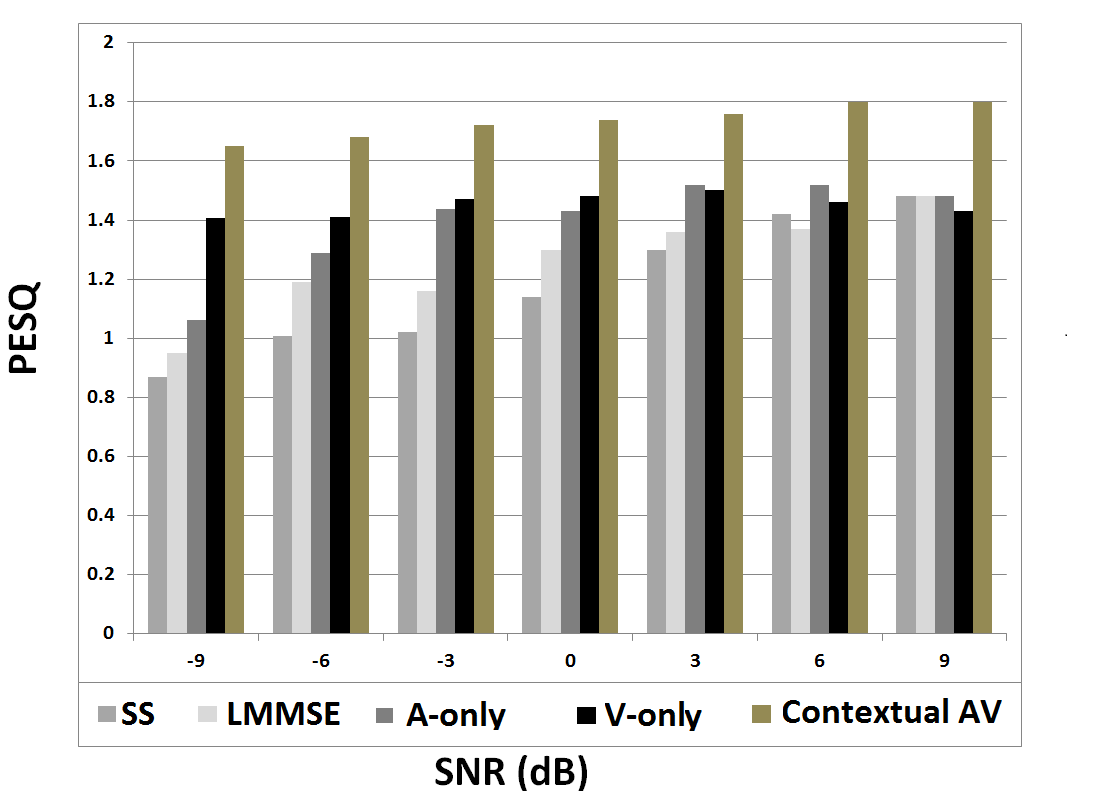}
	\caption{PESQ results with ChiME3 noises. At both low and high SNR levels, EVWF with contextual AV module significantly outperformed SS, LMMSE, A-only, and V-only speech enhancement methods. V-only outperforms A-only at low SNRs and A-only outperforms V-only at high SNRs}
	\label{fig:PESQComparison}
\end{figure}

\begin{figure}
	\centering 
	\subfloat[Clean]{\includegraphics[width=0.5\textwidth]{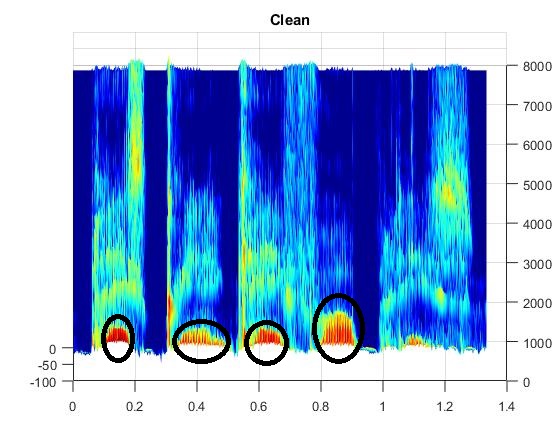}}
	\subfloat[SS enhanced]{\includegraphics[width=0.5\textwidth]{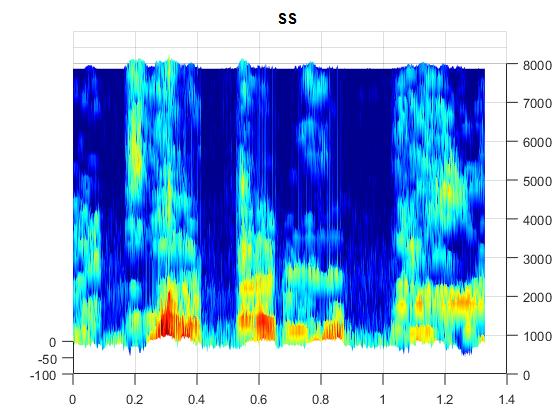}} \\
	\subfloat[Log MMSE Enhanced]{\includegraphics[width=0.5\textwidth]{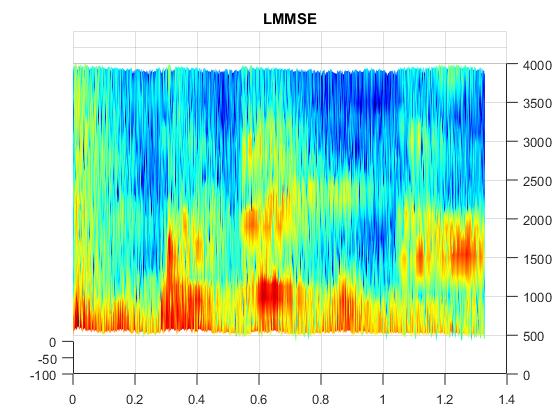}}
	\subfloat[V-Only EVWF Enhanced]{\includegraphics[width=0.5\textwidth]{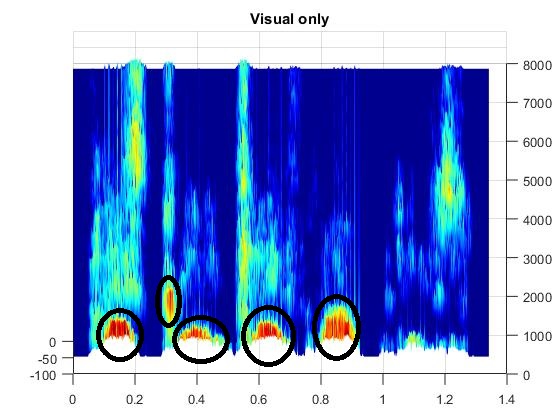}}\\
	\subfloat[A-Only EVWF  Enhanced]{\includegraphics[width=0.5\textwidth]{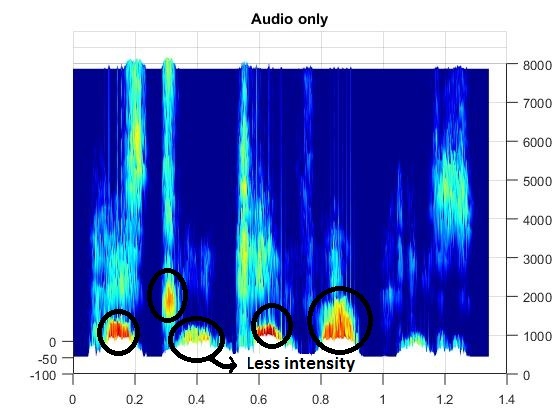}}
	\subfloat[Contextual AV EVWF  Enhanced]{\includegraphics[width=0.5\textwidth]{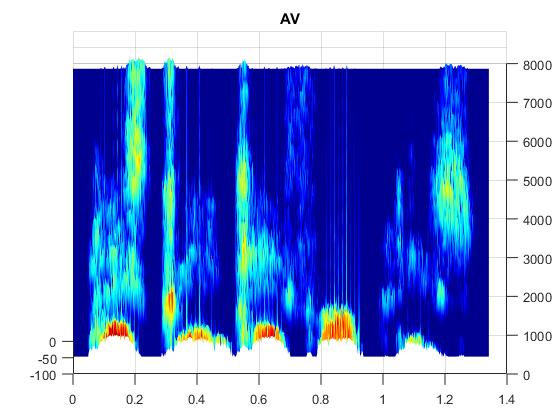}}\\
	\caption{Spectrogram of a randomly selected utterance of -9dB SNR from AV ChiME3 corpus (X-axis: Time; Y-axis: Frequency (Hz)): (a) Clean (b) Spectral Subtraction Enhanced (c) Log MMSE Enhanced (d) Visual-only EVWF (e) Audio-only EVWF (f) Contextual AV EVWF. Note that EVWF with A-only, V-only, and AV outperformed SS and LMMSE. In addition, V-only recovered some of the frequency components better than A-only at low SNR}
	\label{fig:spectrogramComparison}
\end{figure}

\begin{figure}
	\centering
	\includegraphics[trim=0cm 0cm 0cm 0cm, clip=true, width=0.60\textwidth]{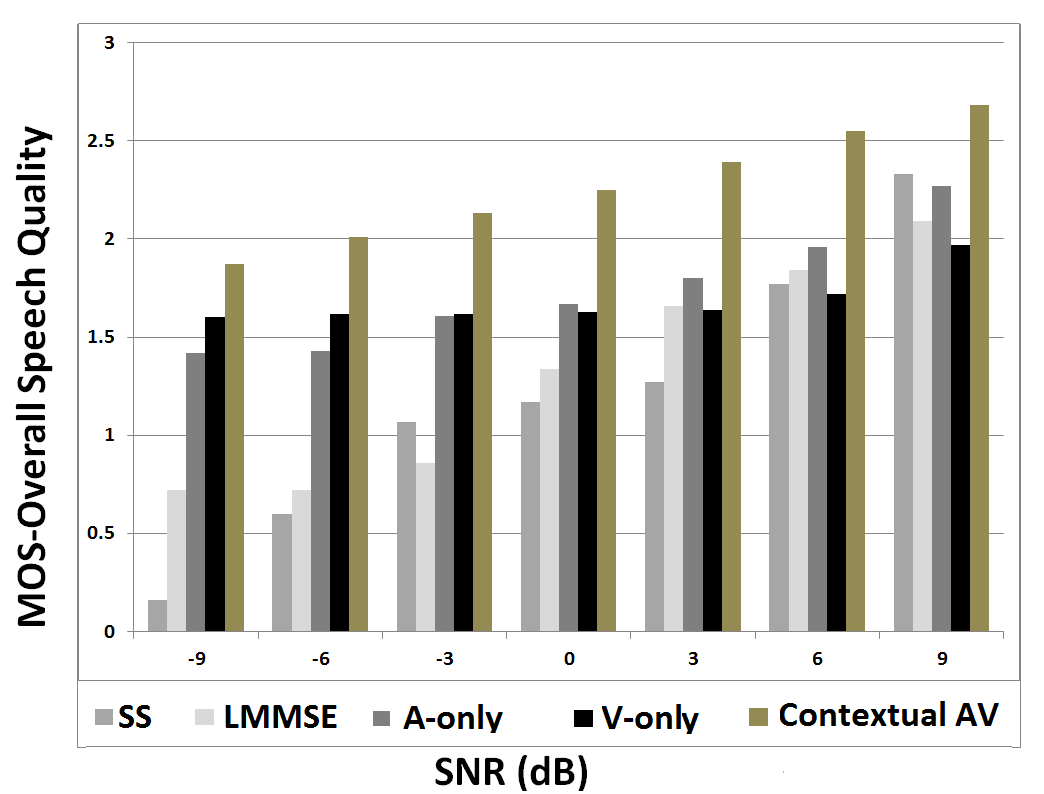}
	\caption{MOS for overall speech quality with ChiME3 noises. Similar to the PESQ results, at both low and high SNR levels, EVWF with contextual AV module significantly outperformed SS, LMMSE, A-only, and V-only speech enhancement methods. It is also to be noted that V-only outperforms A-only at low SNRs and A-only outperforms V-only at high SNRs. However, both A-only and V-only outperform SS and LMMSE at low SNRs and perform comparably at high SNRs.}
	\label{fig:MOSComparison}
\end{figure}

\subsection{Methodology}
The AV switching module was trained over a wide-range of noisy AV ChiME3 corpus, ranging from -9dB to 9dB. A subset of dataset was used for neural network training (70\%) and remaining 30\% was used for network testing (20\%) and validation (10\%). Table 2 summarises the Train, Test, and Validation sentences. For generalisation (SNR independent) testing, the proposed framework was trained on SNRs ranging from -9 to 6dB, and tested on 9dB SNR. 
\subsection{Training and Testing Results}
In this subsection, the training and testing performances of the proposed contextual AV  model are compared with the A-only and V-only models. For preliminary data analysis, V-only model was first trained and tested with different number of prior visual frames, ranging from 1, 2, 4, 8, to 14, and 18 visual frames. The MSE results are depicted in Table 3. It can be seen that by moving from 1 visual frame to 18 visual frames, a significant performance improvement could be achieved. The LSTM model with 1 visual frame achieved the MSE of 0.092, whereas with 18 visual frames, the model achieved the least MSE of 0.058. The LSTM based learning model exploited the temporal information (i.e. prior visual frames) effectively and showed consistent reduction in MSE while going from 1 to 18 visual frames. This is mainly because of its inherent recurrent architectural property and the ability of retaining state over long time spans by using cell gates. The validation results of A-only (A2A), V-only (V2A), and AV (AV2A) are presented in Figure \ref{fig:MSEComparison}. Please note that all three models were trained with a prior 14 frames dataset (an optimal number of prior visual frames for lip movements temporal correlation exploitation). It can be seen that AV contextual model outperformed both A-only and V-only models in achieving least MSE.

\subsection{Speech Enhancement}
\subsubsection{Objective Testing}
For objective testing, PESQ test is used for restored speech quality evaluation. The PESQ score ranges from -0.5 to 4.5 representing least and best possible reconstructed speech quality. The PESQ scores for EVWF (with A-only, V-only, and contextual AV mapping), SS, and LMMSE for different SNRs are depicted in Figure \ref{fig:PESQComparison}. It can be seen that, at low SNR levels, EVWF with AV, A-only, and V-only significantly outperformed SS \cite{boll1979spectral} and LMMSE \cite{ephraim1985speech} based speech enhancement methods. 
At high SNR, A-only and V-only performed comparably to SS and LMMSE methods. It is also to be noted that at low SNR, V-only outperforms A-only and at high SNR, A-only outperforms V-only, validating the phenomena of less effective visual cues at high SNR and less effective audio cues at low SNR. In contrast, the contextual AV EVWF significantly outperformed SS and LMMSE based speech enhancement methods both at low and high SNRs. The low PESQ scores  with ChiME3 corpus, particularly at high SNRs, can be attributed to the nature of the ChiMe3 noise. The latter is characterised by spectro-temporal variation, potentially reducing the ability of enhancement algorithms to restore the signal. However, the AV contextual switching component dealt better with  the spectro-temporal variations. Figure \ref{fig:spectrogramComparison} presents the spectrogram of the clean, SS, LMMSE, V-only, A-only, and contextual AV enhanced audio signals. It can be seen that EVWF with A-only, V-only, and AV outperformed SS and LMMSE at low SNR (-9db). Apparently, there is not much difference in the spectrogram of A-only, V-only, and AV. However, a closer inspection reveals that the V-only method approximating some of the frequency components better than A-only. The energy at specific time-frequency units are highlighted in the snazzy 3-D color spectrogram. 
\subsubsection{Subjective Listening Tests} 
To examine the effectiveness of the proposed contextual AV speech enhancement algorithm, subjective listening tests were conducted in terms of MOS with self-reported normal-hearing listeners using AV ChiME3 corpus. The listeners were presented with a single stimulus (i.e. enhanced speech only) and were asked to rate the re-constructed speech on a scale of 1 to 5. The five rating choices were: (5) Excellent (when the listener feels unnoticeable difference compared to the target clean speech) (4) Good (perceptible but not annoying) (3) Fair (slightly annoying) (2) Poor (annoying), and (1) Bad (very annoying). The EVWF with contextual AV module was compared with A-only, V-only, SS, and LMMSE. A total of 15 listeners took part in the evaluation session. The clean speech signal was corrupted with ChiME3 noises (at SNRs of -9dB, -6dB, -3dB, 0dB, 3dB, 6dB, and 9dB). Figure \ref{fig:MOSComparison} depicts the performances of five different speech enhancement methods in terms of MOS with ChiME3 noises. Similar to the PESQ results, it can be seen that, at low SNR levels, EVWF with contextual AV, A-only, and V-only models significantly outperformed SS and LMMSE methods. At high SNRs, A-only and V-only performed comparably to SS and LMMSE methods. At low SNRs, V-only outperforms A-only and at high SNRs, A-only performs better than V-only, validating the phenomena of less effective visual cues at high SNR and less effective audio cues at low SNR. However, EVWF with contextual AV model significantly outperformed SS, LMMSE, A-only, and V-only methods even at high SNRs exploiting the complementary strengths of both audio and visual cues contextually.  
\section{Conclusion}
In this paper, a novel AV switching component is presented that contextually utilises both visual and noisy audio features to approximate clean audio features in different noise exposures. The performance of our proposed contextual AV switch is investigated using novel EVWF and compared with A-only and V-only EVWF. The EVWF ability to reconstruct clean speech signal is directly proportional to AV mapping accuracy (i.e clean audio features estimation),  better the mapping accuracy, more optimal Wiener filter can be designed for high quality speech reconstruction. Experimental results revealed that the proposed contextual AV switching module successfully exploited both audio and visual cues to better estimate the clean audio features in a variety of noisy environments. The EVWF with contextual AV model outperformed A-only EVWF, V-only EVWF, SS, and LMMSE based speech enhancement methods in terms of both speech quality and intelligibility at both low and high SNRs. At low SNRs, A-only and V-only outperformed SS and LMMSE but performed comparably at high SNRs. In addition, at low SNRs, V-only outperformed A-only and at high SNRs, A-only outperformed V-only. It validates the phenomena of less effective visual cues at high SNRs and less effective audio cues at low SNRs. However, the EVWF with contextual AV model better tackled the spectro-temporal variation at both low and high SNRS. Interestingly, in the snazzy 3-D colour spectrogram results, it is observed that visual cues helped approximating some of the frequency components better than A-only, where larger intensities at specific time-frequency are apparent. In future, we intend to further investigate the performance of our proposed contextual AV module and EVWF in real-world environments with speaker independent scenarios. Though, for the proof of concept, the current study fairly satisfies the speaker independence criteria where a subset of 5 diverse subjects (two white females, two white males, and one black male) with a total of 900 command sentences per speaker was considered. 
\section*{Acknowledgment}
This work was supported by the UK Engineering and Physical Sciences Research Council (EPSRC) Grant No. EP/M026981/1. The authors would like to acknowledge Dr Andrew Abel from Xian Jiaotong-Liverpool University for contribution in audio-visual feature extraction. The authors would also like to acknowledge Ricard Marxer and Jon Barker from the University of Sheffield, Roger Watt from the University of Stirling, and Peter Derleth from Sonova, AG, Staefa, Switzerland for their contributions. The authors would also like to thank Kia Dashtipour for conducting MOS test. Lastly, We gratefully acknowledge the support of NVIDIA Corporation with the donation of the Titan Xp GPU used for this research.

\bibliographystyle{IEEEtran}
\bibliography{referdif1}
\end{document}